\documentclass[lettersize,journal]{IEEEtran}
\usepackage{amsmath,amsfonts}
\usepackage{algorithmic}
\usepackage{algorithm}
\usepackage{array}
\usepackage[caption=false,font=normalsize,labelfont=sf,textfont=sf]{subfig}
\usepackage{textcomp}
\usepackage{stfloats}
\usepackage{url}
\usepackage{verbatim}
\usepackage{graphicx}
\usepackage{cite}
\usepackage{color}
\begin{document}

\title{PointSFDA: Source-free Domain Adaptation for Point Cloud Completion}

\author{Xing He, Zhe Zhu, Liangling Nan, Honghua Chen, Jing Qin, Mingqiang Wei,~\IEEEmembership{Senior Member,~IEEE}
\thanks{X. He, Z. Zhu, H. Chen and M. Wei are with the School of Computer Science and Technology, Nanjing University of Aeronautics and Astronautics, Nanjing, China (e-mail: hexing@nuaa.edu.cn; zhuzhe0619@nuaa.edu.cn; chenhonghuacn@gmail.com; mingqiang.wei@gmail.com).} %
\thanks{L. Nan is with the Urban Data Science Section, Delft University of Technology, Delft, Netherlands (e-mail: liangliang.nan@tudelft.nl).}
\thanks{J. Qin is with the School of Nursing, The Hong Kong Polytechnic University, Hong Kong, China (e-mail: harry.qin@polyu.edu.hk).}
}

\markboth{Journal of \LaTeX\ Class Files,~Vol.~14, No.~8, August~2021}%
{Shell \MakeLowercase{\textit{et al.}}: A Sample Article Using IEEEtran.cls for IEEE Journals}


\maketitle

\begin{abstract}
Conventional methods for point cloud completion, typically trained on synthetic datasets, face significant challenges when applied to out-of-distribution real-world scans. In this paper, we propose an effective yet simple source-free domain adaptation framework for point cloud completion, termed \textbf{PointSFDA}.
Unlike unsupervised domain adaptation that reduces the domain gap by directly leveraging labeled source data, PointSFDA uses only a pretrained source model and unlabeled target data for adaptation, avoiding the need for inaccessible source data in practical scenarios.
Being the first source-free domain adaptation architecture for point cloud completion, our method offers two core contributions.
First, we introduce a coarse-to-fine distillation solution to explicitly transfer the global geometry knowledge learned from the source dataset.
Second, as noise may be introduced due to domain gaps, we propose a self-supervised partial-mask consistency training strategy to learn local geometry information in the target domain. 
Extensive experiments have validated that our method significantly improves the performance of state-of-the-art networks in cross-domain shape completion.
Our code is available at \emph{\textcolor{magenta}{https://github.com/Starak-x/PointSFDA}}.
\end{abstract}


\section{Introduction}
Point clouds are a commonly used 3D data representation in various vision and graphic applications.
However, factors like self-occlusion, light reflection, and limited sensor resolution often result in incomplete and noisy raw-captured point clouds. Therefore, reconstructing the complete shape from its partial observation becomes crucial for downstream tasks.

In recent years, learning-based point cloud completion methods have made remarkable advancements~\cite{PCN,PFNet,CRN,SnowflakeNet,AdaPoinTr,SVDFormer,ProxyFormer}.
These networks are typically trained on rich and clean synthetic datasets, where partial point clouds are generated by virtual scanning~\cite{PCN,AdaPoinTr,CRN}.
Despite the notable progress, transferring the inference capabilities of these models to real-world scans remains challenging due to substantial domain gaps.
To tackle this problem, various techniques have been proposed, including training on unpaired data~\cite{Cycle4completion,pcl2pcl,ShapeInversion} or employing self-supervised learning~\cite{Inpaiting,ACLSPC,P2C}. Due to the lack of direct supervision, their performance is still far from satisfactory. 
Unsupervised domain adaptation (UDA) approaches~\cite{OptDE,SCoDA} seek to bridge the domain gaps by utilizing both source and target datasets for adaptation. 
While in certain critical circumstances such as autonomous driving, the source datasets may be proprietary or commercial~\cite{SFDA4SS}, posing a significant challenge to current UDA methods. 

\begin{figure*}[tbp]
    \centering
    \includegraphics[width=\textwidth]{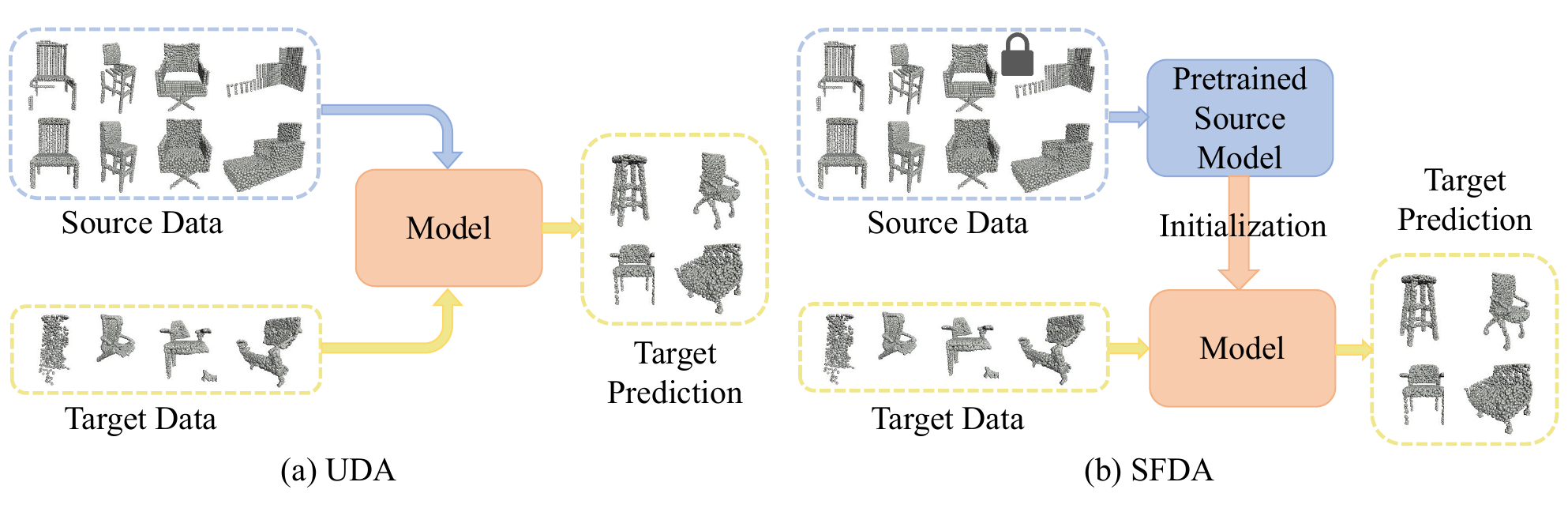}
    \caption{Illustration of (a) unsupervised domain adaptation (UDA) and (b) source-free domain adaptation (SFDA) in point cloud completion. The UDA methods rely on both source and target datasets during adaptation, while SFDA methods only have access to a pretrained source model and unlabeled target data. Both UDA and SFDA aim to generate complete point clouds that resemble the distribution of real data.}
    \label{fig:UDA_SFDA}
\end{figure*}

To address these challenges, we study the problem of source-free domain adaptation (SFDA) for point cloud completion for the first time. SFDA stands out by conducting adaptation exclusively with a pretrained source model and an unlabeled target dataset. Unlike traditional UDA methods, SFDA faces the unique challenge of directly transferring and preserving knowledge from the pretrained model. Considering this approach inherently inevitably introduces noisy information, the key to the SFDA lies in the dual objective of absorbing knowledge from the source model and mitigating the impact of noise during the adaptation process. To achieve this goal, we build a novel framework named PointSFDA, comprising two core components to achieve source knowledge transfer and target information learning, respectively.

To transfer knowledge from the source model, a common strategy in SFDA methods is to distill the source information with the pseudo-label generated by the source model~\cite{SFDAImageDehaze,C-SDFA,Adverseweather,ModelDistillation,NoisyLabelLearning}. However, the performance of the source model, particularly concerning the quality of local details, is highly sensitive to domain shifts, leading to undesired pseudo-labels.
We observe that shapes across different domains exhibit similar global structures but diverge in local point distribution. To this end, we propose Coarse-to-fine Point Cloud Distillation, a method that focuses on transferring domain-invariant global geometry knowledge, minimizing the impact of noise.
In addition, we introduce a Partial-mask Consistency Training strategy to learn the target-relevant information. By ensuring point-wise predictive consistency across various mask augmentations, our method learns the local point distribution of the target domain in a self-supervised manner. 

To evaluate the effectiveness of our method, we conduct experiments in both real-world datasets, including KITTI~\cite{KITTI} and ScanNet~\cite{ScanNet}, and synthetic datasets~\cite{3d-future,ModelNet40}. Experimental results demonstrate that despite the simplicity of our approach, it is highly effective. It significantly enhances the performance of state-of-the-art point cloud completion networks in cross-domain data, outperforming existing UDA and unsupervised methods by a considerable margin.

In summary, our main contributions are as follows:
\begin{itemize}
\item We design the first source-free domain adaptation framework specifically tailored for point cloud completion, making a significant advancement in cross-domain point cloud completion;
\item We propose coarse-to-fine point cloud distillation to effectively transfer geometry knowledge from the pretrained source model to the target model;
\item We introduce partial-mask consistency training, which enables the learning of local geometry information from the target domain in a self-supervised manner and enhances the source model to reduce noise.
\end{itemize}
\label{sec:intro}

\section{Related work}
\subsection{Point Cloud Completion}
Recent works have made impressive progress in point cloud completion in the realm of deep learning~\cite{PCN,PFNet,Morphing,PoinTr,ProxyFormer,AdaPoinTr, CRN,SnowflakeNet,SeedFormer}. 
The pioneering work, PCN~\cite{PCN}, attempts this task with an encoder-decoder network architecture. Specifically, a coarse completion is first produced and then refined by a folding operation~\cite{FoldingNet}. Most following works~\cite{PFNet,Morphing,PoinTr,ProxyFormer,AdaPoinTr,CRN,SnowflakeNet,SeedFormer,Geometric,SVDFormer} adopt a similar coarse-to-fine generation manner.
PoinTr~\cite{PoinTr} formulates point cloud completion as a set-to-set translation problem and employs a transformer encoder-decoder architecture. SnowflakeNet~\cite{SnowflakeNet} designs the decoder so that the process of generating points resembles the growth of snowflakes, aiming to produce intricate details. 
There are also supervised methods that model point cloud completion as a deformation task. PMP-Net~\cite{PMPNet} and PMP-Net++\cite{PMPNet++} move each point from the partial input to achieve a complete point cloud, ensuring that the total distance traveled along the moving paths (PMPs) is minimized. 
Additionally, there exists a line of works~\cite{ViPC,XMFNet,CSDN} that utilize additional monocular images to assist point cloud completion. 
Although these supervised methods perform exceptionally well on synthetic datasets, they typically struggle to generate highly accurate completions when handling cross-domain real-world data. 

Unsupervised point cloud completion methods use unpaired partial and complete point clouds for training. GAN-based methods~\cite{pcl2pcl,ShapeInversion,Cycle4completion} tackle this problem by adversarial training. Pcl2pcl~\cite{pcl2pcl} pretrains two auto-encoders for feature extraction and employs a generator to generate complete features from partial point clouds. A discriminator is then used to distinguish whether the features originate from complete or partial point clouds. ShapeInversion~\cite{ShapeInversion} use a pretrained generator to generate complete point clouds from partial ones, then utilizes a degradation function to degrade the complete point clouds into partial ones. The discriminator assesses the feature distance between the degraded partial point clouds and the input partial point clouds. Cycle4Completion~\cite{Cycle4completion} employs CycleGAN~\cite{CycleGAN} to simultaneously learn cycle transformations from the completion domain to the incomplete domain and from the incomplete domain to the completion domain. 
Although these methods do not require paired data, the output point clouds conform to the distribution of the synthetic data, leading to a domain gap with real-world data. 

Self-supervised point cloud completion methods take a step further and train a network with only partial point clouds.
Some approaches~\cite{Inpaiting,P2C} remove a portion of the input data and then recover the missing part of an object by observing similar regions of other objects in the same category.
ACL-SPC~\cite{ACLSPC} employs an adaptive closed-loop (ACL) system to ensure consistent output despite variations in input. While these methods alleviate the need for complete point clouds, the resulting complete point clouds often exhibit the average shapes of the training categories.

To address the aforementioned issues, one potential solution is to employ Unsupervised Domain Adaptation (UDA). 
OptDE~\cite{OptDE} disentangles partial point cloud to occlusion factor, domain factor, and shape factors.
The completion is achieved by transferring domain-invariant shape features from the source domain to the target domain and ignoring the occlusion. 
However, in certain critical circumstances such as autonomous driving, the source datasets may be proprietary and commercial, limiting access to only the source models and unlabeled target datasets.

\subsection{Source-free Domain Adaptation}
Source-free domain adaptation methods aim to transfer a pretrained source model to the target domain without access to the source datasets. 

The primary concern for SFDA is effectively transferring knowledge from the source domain to the target domain. 
Some methods~\cite{SoFa,SFDA4SS,DomainImpression,AvatarPrototype} generate synthetic source domain data to facilitate the transfer of domain knowledge from a well-trained source model to a target model. Other methods~\cite{SFDAImageDehaze,SFDAVideo} design straightforward modules to align features between the source and target domains. Despite variations in methodology, the transfer of domain knowledge from the source to the target domain typically occurs in the form of features and may not encompass the transfer of geometric information.

In addition, effectively learning target domain knowledge is also crucial in SFDA. SHOT~\cite{SHOT} learn the target-specific feature encoding module using self-supervised learning and semi-supervised learning, with the source hypothesis fixed. SFDAHPE~\cite{SFDAHPE} mitigates catastrophic forgetting of the source domain knowledge through the use of an intermediate model. NRC~\cite{Neighborhood} leverages the intrinsic clustering of target data by nearest neighbor.  Hegde et al. \cite{Adverseweather} use mean teacher with Monte-Carlo dropout to reduce the uncertainty of noisy labels. AUGCO~\cite{AUGCO} adopts Augmentation Consistency-guided Self-training, selecting only reliable samples for training through a confidence threshold. C-SFDA~\cite{C-SDFA} uses curriculum learning-aided self-training to select noisy labels, while not requiring a memory bank like previous curriculum learning methods. However, curriculum learning or selective pesuo label requires differences between samples (or versions of data augmentation) to measure the label reliability, while in most categories of point cloud completion tasks, there are no significant differences between samples.

However, directly applying the aforementioned domain adaptation methods to point cloud completion may not yield satisfactory results, as point cloud completion involves a point-wise generation task. Therefore, our method distills point-wise coordinates and learns target information in a self-supervised manner, rather than relying on features or pseudo-labels.


\label{sec:related}

\section{Method} 
\subsection{Problem Description and Overview}
In the scenario of source-free domain adaptation, we are given a point cloud completion network $\Phi^s$ trained on a source dataset. 
Our goal is to train a target model $\Phi^t$ with only access to the source pretrained model $\Phi^s$ and the target dataset $\left\{ {P}_{in}^{i,t}  \right\} _{i=1}^{N}$, which compromises $N$ partial point clouds without ground truth. 
We masked point clouds$\left\{ P_{in,i}^{t}\right\}^{K}_{i=0}$, where $K$ represents the number of masked point clouds in partial-mask consistency training, and $P_{in,0}^{t}$ equals the original partial point cloud $P_{in}^{t}$. In our setting, the target model $\Phi^t$ shares the same architecture with $\Phi^s$. Following previous point cloud completion methods~\cite{PCN,PFNet,Morphing,PoinTr,ProxyFormer,AdaPoinTr,CRN,SnowflakeNet,SeedFormer,Geometric,SVDFormer}, we assume the network generates results in a coarse-to-fine manner, where a coarse shape ${P}_{coarse}^{s}$ ($\left\{{P}_{coarse,i}^{t}\right\}^{K}_{i=0}$) is first produced and then refined to the final result ${P}_{fine}^{s}$ ($\left\{{P}_{fine,i}^{t}\right\}^{K}_{i=0}$).

\begin{figure*}[h]
    \centering
    \includegraphics[width=\textwidth]{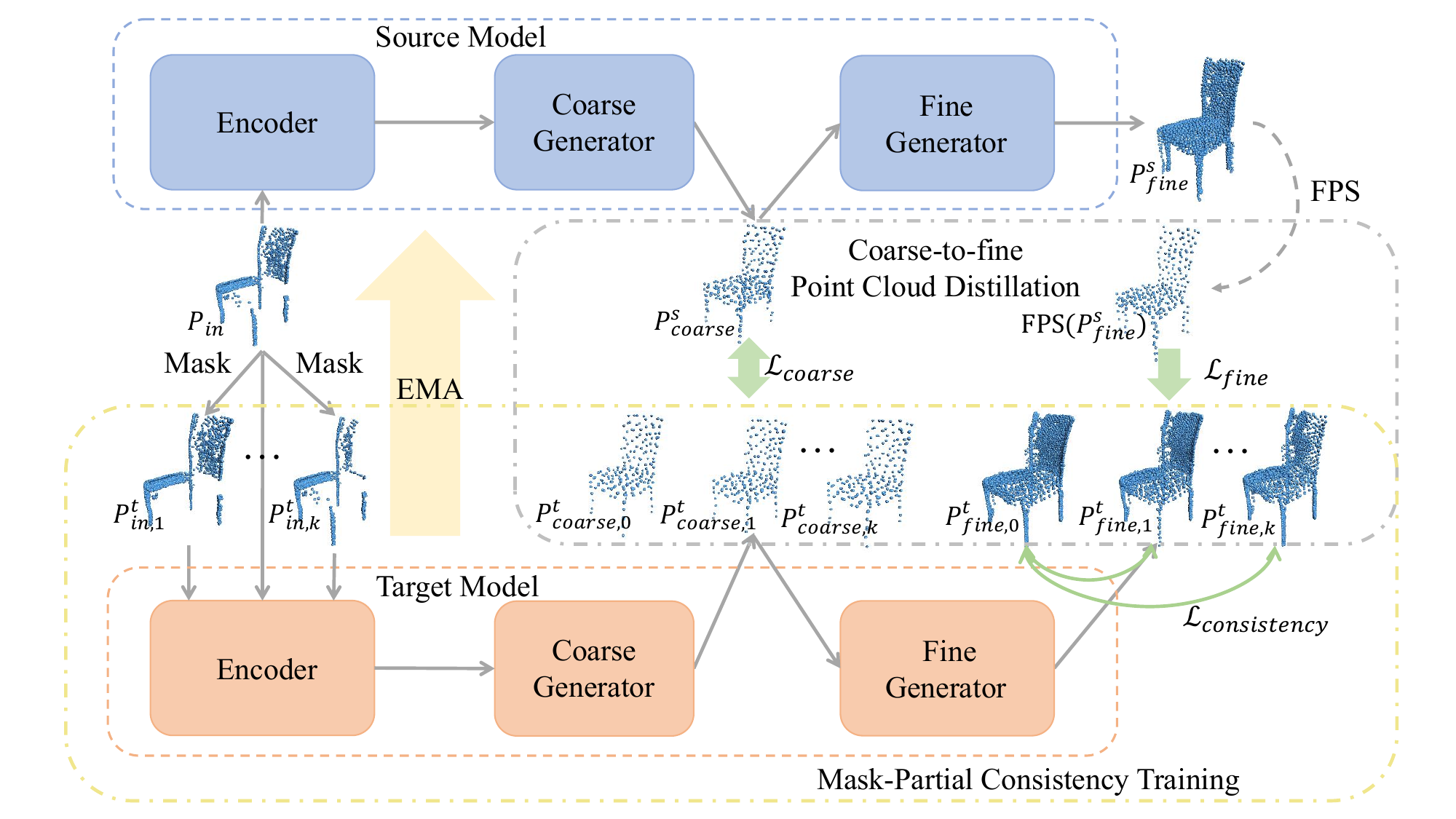}
    \caption{Overview of the proposed PointSFDA. 
    A partial point cloud from the target domain is directly used as input for the source model and masked $k$ times ($k = 2$ in this figure) before being fed into the target model. Coarse-to-fine Point Cloud Distillation is employed in the coarse point cloud and fine point cloud to directly transfer the geometry information across domains. Mask-partial Consistency Training learns target data geometry information through point-wise predictive consistency across various mask augmentations.}
    \label{fig:pipeline}
\end{figure*}

The overall framework of PointSFDA is illustrated in Figure~\ref{fig:pipeline}. Our approach comprises two core components: Coarse-to-fine Point Cloud Distillation and Partial-Mask Consistency Training. The former serves as a knowledge transfer module to exploit the learned domain-invariant information within $\Phi^s$,
while the latter focuses on learning the target-relevant information.

\subsection{Coarse-to-fine Point Cloud Distillation}
Due to the absence of source data, existing SFDA methods often resort to employing knowledge distillation techniques to transfer knowledge learned from rich source datasets~\cite{SFDAImageDehaze,C-SDFA,Adverseweather,ModelDistillation,NoisyLabelLearning}. 
Taking target data as input, these methods use the output of the source model as a pseudo-label to supervise the target model. Nevertheless, the completion quality of local details significantly decreases when the input establishes large domain gaps. Thus, a mere adaptation of vanilla knowledge distillation to our task will introduce a significant amount of noise into the target model. 
Besides, we observe that the global structure of a point cloud within a particular category remains domain-invariant. Based on these insights, we propose a coarse-to-fine distillation approach, which transfers only global-level geometry knowledge to both the coarse completion and final result of the target domain.


For fine point clouds, we first utilize Farthest Point Sampling (FPS) to downsample the output point cloud. This step aims to capture the global structure of the fine point cloud and mitigate noise. Then, we calculate the unidirectional chamfer distance from the sampled point cloud to the output point cloud of the target model, i.e.,
\begin{equation}
\label{eq:fine}
  \mathcal{L}_{fine}=\sum_{i=0}^{K}UCD\left( {FPS}\left({P}_{fine}^{s}\right),{P}_{fine,i}^{t}\right),
\end{equation}
where the Unidirectional Chamfer Distance (UCD) is given by
\begin{equation}
  {UCD}\left( X,Y\right)=\frac{1}{X}\sum_{x\in X} \mathop{\min}_{y \in Y} \| x - y \|, 
\end{equation}
with $X$ and $Y$ denoting two point clouds.

For coarse point clouds, distillation enables the target model to better learn the global structure of point clouds during the training stage. We use chamfer distance to minimize the discrepancy between the coarse point clouds generated by the source model ${P}_{coarse}^{s}$ and those produced by the target model ${P}_{coarse}^{t}$,
\begin{equation}
\label{eq:coarse}
\mathcal{L}_{coarse}=\sum_{i=0}^{K}CD\left( {P}_{coarse}^{s},{P}_{coarse,i}^{t}\right),
\end{equation}
where the Chamfer Distance (CD) is given by
\begin{equation}
  {CD}\left( X,Y\right)=\frac{1}{X}\sum_{x\in X} \mathop{\min}_{y \in Y} \| x - y \| +\frac{1}{Y} \sum_{y\in Y} \mathop{\min}_{x \in X} \| y - x \|,
\end{equation}
with $X$ and $Y$ denoting two point clouds.

\subsection{Partial-Mask Consistency Training}
With the global knowledge transferred from the pretrained source model, the target model has learned the overall structure of the input from the target domain. However, the reconstructed local structures are still undesired without learning the target-relevant knowledge. 

To tackle this issue, we propose Partial-Mask Consistency Training (PMCT), which aims to learn the local geometry information of the target data by ensuring consistency in point-wise predictions across different mask augmentations. Initially, inspired by the masking methodology in~\cite{Inpaiting}, we partition the input point cloud $P_{in}$ into eight equal parts in the 3D space and randomly mask one of them each time to obtain $\left\{ P_{in,i}^{t}\right\}^{K}_{i=0}$. Then, we feed $\left\{ P_{in,i}^{t}\right\}^{K}_{i=0}$ into the target model and apply the coarse-to-fine distillation to all of them. Ultimately, we obtain the output point clouds $\left\{ P_{fine,i}^{t}\right\}^{K}_{i=0}$ and use chamfer distance to make the output of each masked point cloud as consistent as possible with $P_{fine,0}^{t}$, i.e.,
\begin{equation}
\label{eq:consistency}
\mathcal{L}_{consistency}=\sum_{i=1}^{K}CD\left( P_{fine,0}^{t}, P_{fine,i}^{t}\right).
\end{equation}


With the masking operations described above, different local geometric structures of objects may be discarded, leading to variations in the masked regions of the same input point clouds. By computing consistency among these regions, our method  strengthens the model's perception of local geometry.


To leverage the knowledge acquired by the target model to rectify the source model, we follow~\cite{Meanteacher,MutualMeanteacher} to employ Exponential Moving Average (EMA) to correct the parameters of the source model:
\begin{equation}
\label{eq:ema}
    \Phi^s = \eta \Phi^s +\left(1-\eta \right) \Phi^t 
\end{equation}
where $ \eta \in \left[0, 1\right]$ represents the decay rate, a hyper-parameter that controls the update rate of the source model.


\subsection{Overall loss function}
In addition to the loss functions mentioned above given by Equations~\eqref{eq:fine}, ~\eqref{eq:coarse}, and ~\eqref{eq:consistency}, the final output point cloud needs to preserve the structure of the partial point cloud. To achieve this, we employ the partial match loss, i.e., 
\begin{equation}
\mathcal{L}_{partial}=\sum_{i=0}^{K}UCD\left( {P}_{in},{P}_{fine,i}^{t}\right).
\end{equation}

During the joint training, the overall loss is defined as:
\begin{equation}
  \mathcal{L} =\lambda_1 \mathcal{L}_{fine}+ \lambda_2 \mathcal{L}_{coarse} +\lambda_3 \mathcal{L}_{consistency} + \lambda_4 \mathcal{L}_{partial}
\end{equation}
where $\lambda_1$, $\lambda_2$, $\lambda_3$, and $\lambda_4$ are tradeoff hyperparameters.
\label{sec:method}

\section{Experiments}
\subsection{Implementation Details}
Our method is implemented using PyTorch~\cite{Pytorch} and trained on a single NVIDIA 4090 GPU. 
In the coarse-to-fine point cloud Distillation, The output point cloud ${P}_{fine}^{s}$ from the source model is downsampled to 256 points. 
In the Self-supervised Partial-Mask Consistency, we set $K$ to 1. $\lambda_1$, $\lambda_2$, $\lambda_3$ and $\lambda_4$ are set as $1$, $1$, $1e2$, $1e2$, respectively.
We apply the proposed PointSFDA to three representative point cloud completion networks: PCN~\cite{PCN}, SnowflakeNet~\cite{SnowflakeNet}, and AdaPoinTr~\cite{AdaPoinTr}. In all the experiments, the input and output point clouds are downsampled to 2,048 points.


\subsection{Dataset and Evaluation Metrics}
Following OptDE~\cite{OptDE}, we use the CRN~\cite{CRN} dataset as the source domain dataset in all the experiments. CRN is derived from ShapeNet~\cite{ShapeNet},  which consists of eight categories: plane, car, cabinet, chair, lamp, sofa, table, and watercraft, totaling 30,174 partial-complete pairs. We utilize the same partitioning strategy as the OptDE~\cite{OptDE} dataset for training and evaluation.

To evaluate PointSFDA in real-world scans, we first conduct experiments on KITTI~\cite{KITTI} and ScanNet~\cite{ScanNet}. The cars in KITTI and the chairs, lamps, sofas, and tables in ScanNet are used. To facilitate the quantitative evaluation of performance on real datasets, we utilize the ground truth provided by the ScanSalon dataset~\cite{SCoDA}, where the high-quality complete 3d models are created by artists.

In addition, we test PointSFDA on 3D-FUTURE~\cite{3d-future} and ModelNet~\cite{ModelNet40} for evaluation under different kinds of domain shifts.
3D-FUTURE incorporates photo-realistic synthetic images alongside high-resolution textured 3D CAD furniture shapes crafted by professional designers. As a result, samples in 3D-FUTURE closely mimic real-life objects. The partial and complete shapes within 3D-FUTURE are captured from five distinct viewpoints. 
ModelNet is derived from ModelNet40~\cite{ModelNet40}, where The complete shapes are formed from randomly sampled points on the surface, while partial shapes are generated through virtual scanning techniques. The complete and incomplete point clouds in 3DFUTURE and ModelNet are both sampled to 2048 points, to align with CRN.
We employ Chamfer Distance with $L_2$ normalization as the evaluation metric. 

\subsection{Real Data Evaluation}
\begin{table}[h!]
  \renewcommand\arraystretch{1.2}
  \caption{Completion results on Real dataset in terms of per-point $L_2$ Chamfer Distance$\times 10^4$ (lower is better).
  }
  \label{tab:real}
  \centering
  \begin{tabular}{c|cccccc}
    \hline
    Method & Car & Chair & Lamp & Sofa & Table & Avg. \\
    \hline
    ACL-SPC~\cite{ACLSPC} & 31.94 & 20.03 & 35.05 & 31.09 & 25.08 & 28.64\\
    P2C~\cite{P2C}   & 25.47 & 20.30 & 51.59 & 38.08 & 22.47 & 31.58 \\
    OptDE~\cite{OptDE} & 26.89 & 18.18 & 40.62 & 20.09 & 28.15 & 26.79  \\
    \hline
    PCN~\cite{PCN} & 37.09 & 25.33 & 66.46 & 18.14 & 60.43 & 41.49 \\
    SFDA-PCN &34.91 &16.13 & 32.69 & 15.32 & 27.42 & 25.29\\
    AdaPoinTr~\cite{AdaPoinTr} & 34.52 & 19.18 & 70.41 & 15.90 & 46.72 & 37.35 \\
    SFDA-AdaPoinTr & \bf 20.51 & 13.49 & \bf 30.32 & \bf 11.17 & \bf21.85 & \bf19.47\\
    SnowflakeNet~\cite{SnowflakeNet} & 32.39 & 15.69 & 51.00 & 12.45 & 33.53 & 29.01 \\
    SFDA-SnowflakeNet &29.89 & \bf 12.81 & 39.52 & 12.35 & 21.97 & 23.31 \\
  \hline
  \end{tabular}
\end{table}

We conduct experiments on KITTI~\cite{KITTI} and ScanNet~\cite{ScanNet} for real data evaluation. We compare our method with self-supervised methods P2C~\cite{P2C}, ACL-SPC~\cite{ACLSPC}, and UDA method OptDE~\cite{OptDE}.

The quantitative results are summarized in Table~\ref{tab:real}, where PCN, AdaPoinTr, and SnowflakeNet denote these models pretrained on the CRN dataset without adaptation. SFDA-PCN, SFDA-AdaPoinTr, and SFDA-SnowflakeNet represent the corresponding models adapted by the proposed PointSFDA.

It can be observed that, although the source models are well-trained on the synthetic CRN dataset, the performance is far from satisfactory when tested on real-world data, typically performing worse than the self-supervised and UDA methods. After adaptation, all three models have shown significant improvements. In particular, our method can reduce the CD value of AdaPoinTr by 47.9\%. Meanwhile, without access to the source data, all three models achieve superior performance compared to the state-of-the-art UDA method OptDE. This further outlines the ability of PointSFDA to effectively adapt to the target domain data.

\begin{figure*}[t]
    \centering
    \includegraphics[width=\textwidth]{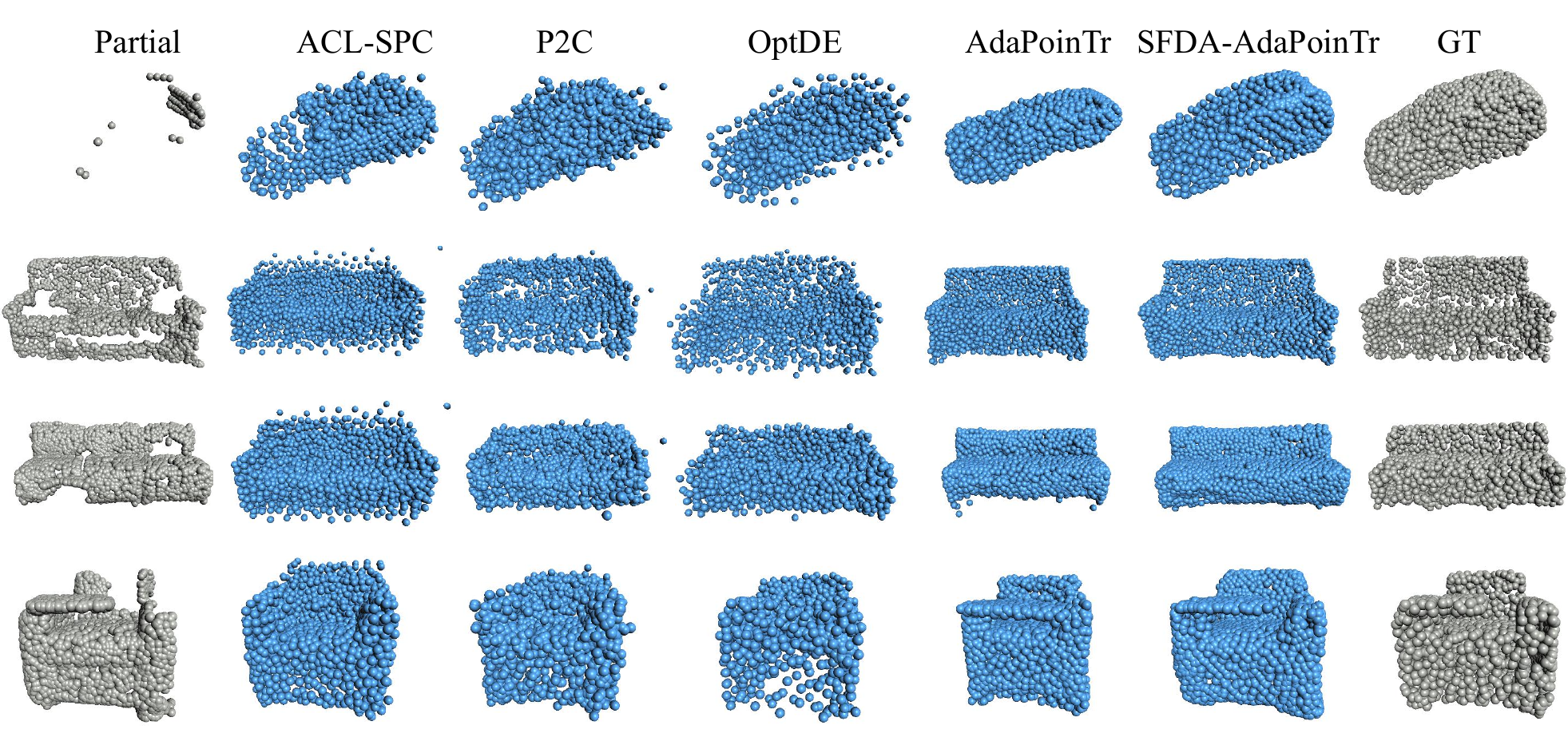}
    \caption{Visual comparison with recent methods(ACL-SPC~\cite{ACLSPC}, P2C~\cite{P2C}, OptDE~\cite{OptDE}, AdaPoinTr~\cite{AdaPoinTr}) on the test set of KITTI and ScanNet.}
    \label{fig:real}
\end{figure*}

In addition, we present a visual comparison in Figure~\ref{fig:real}. 
Notably, when encountering a domain gap, AdaPoinTr exhibits errors in both global structure and geometric details. For example, the armrests of the chair model are merged. With the assistance of PointSFDA, the network avoids this issue and produces a result closer to the ground truth.
Meanwhile, it is evident that our SFDA-AdaPoinTr significantly outperforms ACL-SPC, P2C, and OptDE by a large margin.

\subsection{Synthetic Data Evaluation}
In this study, we compare our method with existing unsupervised point cloud methods, including Pcl2Pcl~\cite{pcl2pcl}, ShapeInversion~\cite{ShapeInversion}, and Cycle4completion~\cite{Cycle4completion}, as well as self-supervised point cloud completion methods including P2C~\cite{P2C}, ACL-SPC~\cite{ACLSPC}, and UDA methods like OptDE~\cite{OptDE}.

The quantitative results for the 3DFUTURE dataset and ModelNet dataset are presented in Table~\ref{tab:3DFUTURE} and Table~\ref{tab:ModelNet}, respectively. It is clear that our methods surpass other approaches by a significant margin. Our outstanding performance across different datasets demonstrates its capacity to adapt to diverse domain gaps. 

\begin{table}[h!]
  \renewcommand\arraystretch{1.2}
  \caption{Completion results on the 3DFuture dataset in terms of per-point $L_2$ Chamfer Distance$\times 10^4$ (lower is better).
  }
  \label{tab:3DFUTURE}
  \centering
  \setlength{\tabcolsep}{4pt} 
  \begin{tabular}{c|cccccc}
     \hline
    Method & Cabinet & Chair & Lamp & Sofa & Table & Avg. \\
     \hline
    Pcl2pcl~\cite{pcl2pcl} 
    & 57.23 & 43.91 & 157.86 & 63.23 & 141.92 & 92.83\\
    ShapeInversion~\cite{ShapeInversion} 
    & 38.54 & 26.30 & 48.57 & 44.02 & 108.60 & 53.21\\
    Cycle4Completion~\cite{Cycle4completion} 
    & 32.62 & 34.08 & 77.19 & 43.05 & 40.00 & 45.39\\
    P2C~\cite{P2C} & 43.01 & 24.57 & 26.52 & 41.60 & 22.69 & 31.68 \\
    ACL-SPC~\cite{ACLSPC} & 70.12 & 23.87 & 31.75 & 28.74& 25.38 & 35.97 \\
    OptDE~\cite{OptDE} & 28.37 & 21.87 & 29.92 & 37.98 & 26.81 & 28.99 \\
     \hline
    PCN~\cite{PCN} & 39.13 & 28.16 & 86.34 & 33.62 & 71.37 & 51.72 \\
    SFDA-PCN &24.73 & 17.94 & 27.54 & 19.60 & 22.43  & 22.45\\
    AdaPoinTr~\cite{AdaPoinTr} & 26.96 & 22.71 & 50.45 & 25.24& 45.26 & 34.12 \\
    SFDA-AdaPoinTr & 16.72 & 12.31 & 17.44 & 14.15 & 14.47 & 15.08 \\
    SnowflakeNet~\cite{SnowflakeNet} & 19.43 & 17.08  & 32.84 & 19.20 & 29.26 & 23.80 \\
    SFDA-SnowflakeNet & \bf15.82 & \bf11.93 & \bf15.04 & \bf13.43 & \bf13.93 & \bf14.04 \\
   \hline
  \end{tabular}
\end{table}


\begin{table*}[h!]
  \renewcommand\arraystretch{1.2}
  \caption{Completion results on ModelNet dataset in terms of per-point $L_2$ Chamfer Distance$\times 10^4$ (lower is better).
  }
  \label{tab:ModelNet}
  \centering
  \begin{tabular}{c|ccccccc}
     \hline
    Method & Plane & Car & Chair & Lamp & Sofa & Table & Avg. \\
     \hline
    Pcl2pcl~\cite{pcl2pcl} &18.53 & 17.54 & 43.58 & 126.80 & 38.78 & 163.62 & 68.14\\
    ShapeInversion~\cite{ShapeInversion} & 3.78 & 15.66 & 22.25 & 60.42 & 22.25 & 125.31 & 41.61\\
    Cycle4Completion~\cite{Cycle4completion} & 5.77 & 11.85 & 26.67 & 83.34 & 22.82 & 21.47 & 28.65\\
    P2C~\cite{P2C} & 4.80 & 19.66 & 17.68 & 44.69 & 32.26 & 12.83 & 21.99 \\
    ACL-SPC~\cite{ACLSPC} & 5.75 & 11.73 & 43.08 & 106.29 & 25.62 & 16.89 & 34.89 \\
    OptDE~\cite{OptDE} & 2.18 & 9.80 & 14.71 & 39.74 & 19.43 & \bf9.75 & 15.94\\
     \hline
    PCN~\cite{PCN} & 5.12 & 9.38 & 25.45 & 75.03 & 19.24 & 81.12 & 35.89 \\
    SFDA-PCN & 2.49 & 9.56 & 14.31 & 48.02 & 17.39 & 17.89 & 18.28 \\
    AdaPoinTr~\cite{AdaPoinTr} & 2.39 & 8.06 & 20.15 & 45.37 & 15.88 & 55.93 & 24.63 \\
    SFDA-AdaPoinTr & 1.99 & 7.80 & \bf9.89 & \bf28.53 & 11.74 & 12.41 & 12.06 \\
    SnowflakeNet~\cite{SnowflakeNet} & 2.50 & 7.93 & 14.55 & 40.95 & 12.79 & 24.44 & 17.19 \\
    SFDA-SnowflakeNet & \bf1.72 & \bf7.59 & 10.43 & 28.60 & \bf10.71 & 10.75 & \bf11.63 \\
   \hline
  \end{tabular}
\end{table*}

\begin{figure*}[h]
    \centering
    \includegraphics[width=\textwidth]{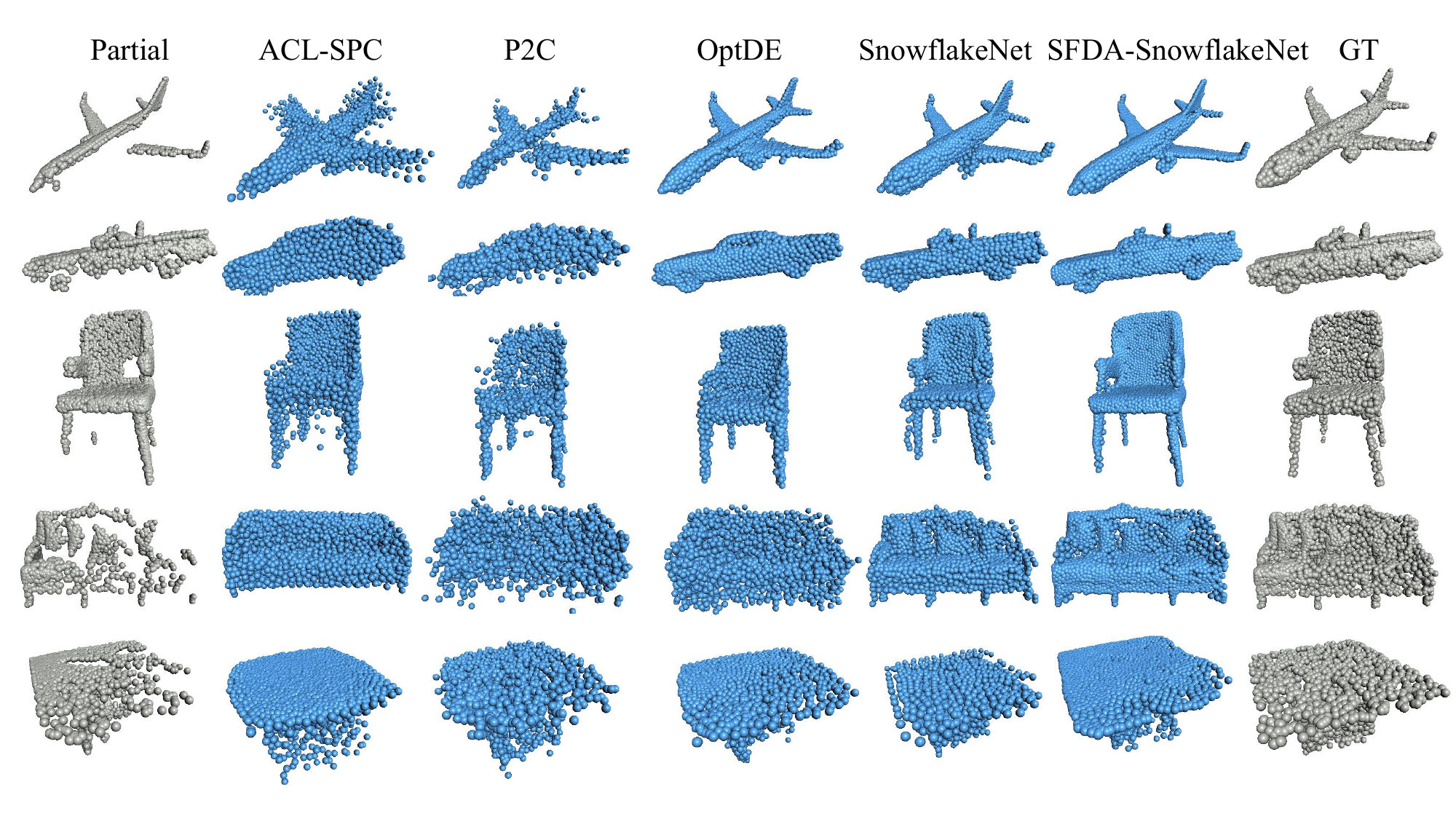}
    \caption{Visual comparison with recent methods(ACL-SPC~\cite{ACLSPC}, P2C~\cite{P2C}, OptDE~\cite{OptDE}, SnowflakeNet~\cite{SnowflakeNet}) on the test set of 3D-FUTURE and ModelNet. The plane and car come from ModelNet, while the chair, sofa, and table are from 3D-FUTURE.}
    \label{fig:systhetic}
\end{figure*}

We present some randomly selected qualitative results of 3DFUTURE and ModelNet in Figure~\ref{fig:systhetic}. Planes and cars are from ModelNet, while chairs, sofas, and tables are from the 3DFUTURE dataset. It is evident that ACL-SPC and P2C tend to generate average shapes. Additionally, P2C tends to produce a significant amount of noise points. While OptDE effectively restores the overall shape, it fails to complete the engine of the airplane and the armrests of the chair.

Due to the domain gap, SnowflakeNet incorrectly completes the overall shape, as seen in the erroneous chair legs and the incorrect size of the tabletop. In contrast, our SFDA-SnowflakeNet completes entire shapes and recovers finer details, outperforming the aforementioned methods.

\subsection{Ablation Study}
In this section, we conduct extensive ablation studies on the key components of our PointSFDA framework.
All experiments in our ablation study utilize SnowflakeNet as the backbone and are conducted on the 3DFUTURE dataset.

\subsubsection{Ablation on Coarse-to-fine Point Cloud Distillation}
We start from the baseline model A, where only the partial match loss $L_{partial}$ and $L_{fine}$ are used. Subsequently, we incorporate $L_{coarse}$ in variant B, which uses coarse-to-fine point cloud distillation without partial-mask consistency training. The improvement indicates that $L_{coarse}$ can effectively transfer the global structure. Variant C only employs partial-mask consistency training in a self-supervised manner. Compared to our method, variant C shows an average increase of 4.65 in ${CDL}_2$, indicating the effectiveness of the coarse-to-fine point cloud distillation.
In Variant D, we replace $L_{coarse}$ with $L_{feature}$, which minimizes the cosine similarity between the global features of the source model and the target model. The performance drop indicates that the distillation of global features can transfer the global structure to some extent, but its effectiveness is not as good as our point cloud distillation.

\subsubsection{Ablation on Partial-Mask Consistency Training}
Additionally, to validate the effectiveness of Partial-Mask Consistency Training (PMCT), we utilize Mean-Teacher (MT) as a replacement in variant E. Our method demonstrates a significant improvement compared to Variant E, with an average increase in $CDL_2$ of 4.59. Meanwhile, comparing Variant E with Variant B, it can be observed that adding a Mean-Teacher to Coarse-to-fine point cloud distillation provides insignificant benefits.

\begin{figure}[h]
\includegraphics[width=0.5\textwidth]{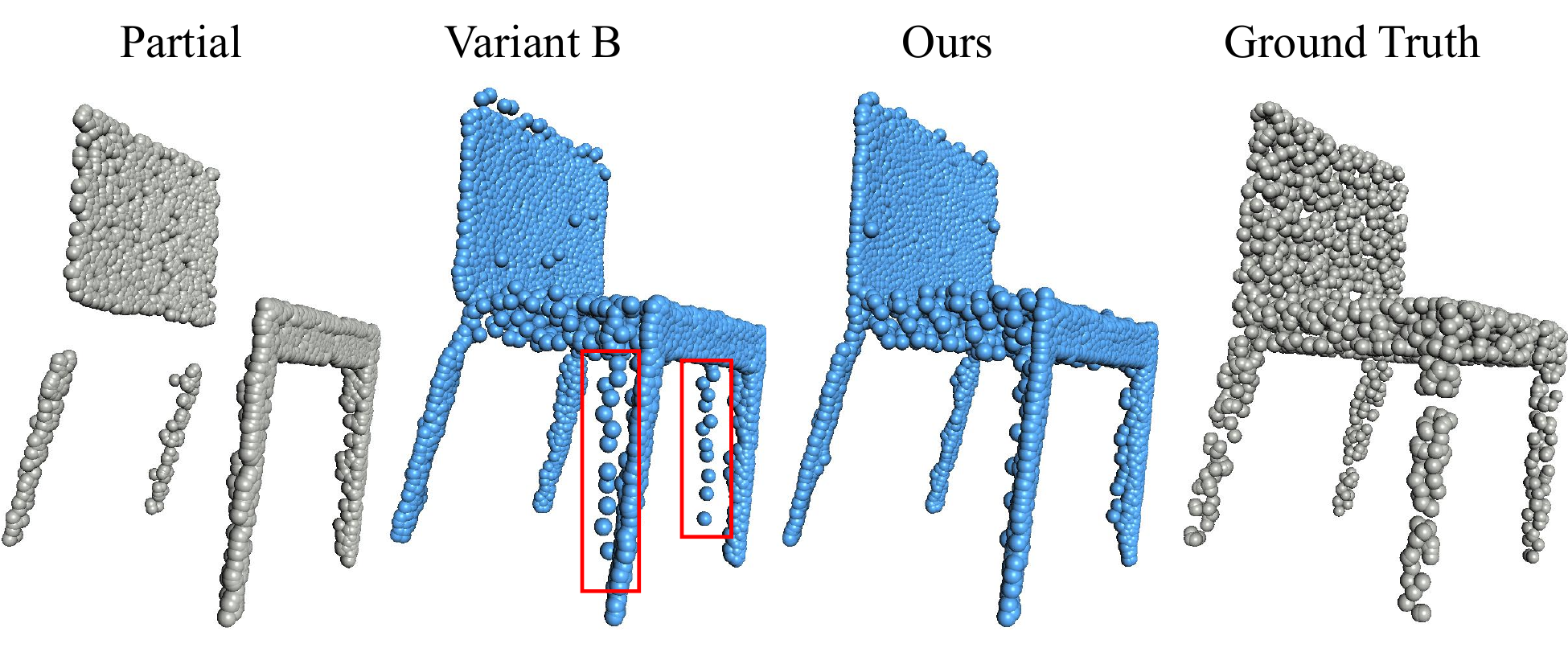}
    \caption{The ablation study of PMCT. In Variant B, only coarse-to-fine point cloud distillation is used, without employing PMCT. Notably, Variant B may recover false chair legs due to the domain gap.}
    \label{fig:problem}
\end{figure}

Moreover, our visualization in Figure~\ref{fig:problem} demonstrates that exclusively employing Coarse-to-fine point cloud distillation leads to inaccuracies in local structures, such as noise points near chair legs, attributed to domain gap. However, the addition of PMCT significantly alleviates this problem.


\begin{table*}[htbp]
  \renewcommand\arraystretch{1.2}
  \caption{Ablation study on 3DFUTURE dataset in terms of per-point $L_2$ Chamfer Distance$\times 10^4$ (lower is better). We utilize SnowflakeNet as the backbone. MT and PMCT refer to Mean-Teacher and Partial-Mask Consistency Training, respectively.
  }
  \label{tab:ablation}
  \centering
  \begin{tabular}{c|ccccc|cccccc}
    \hline
    Variants&$L_{fine}$ & $L_{coarse}$ & $L_{feature}$ &  MT & PMCT
    & Cabinet & Chair & Lamp & Sofa & Table & Avg. \\
    \hline
     A&\checkmark& & & & &  18.19 & 13.64 & 26.08 & 16.83 & 21.04 & 19.16\\
     B&\checkmark& \checkmark& & & & 18.52 & 13.56 & 25.39 & 16.39  & 20.99 & 18.97\\
     C&& & & & \checkmark& 18.48 & 19.74 & 17.79 & 20.46 & 16.99 & 18.69\\
     D&\checkmark&  & \checkmark& & \checkmark & 16.35 & 13.32 & 16.39 & 14.87 & 15.27 & 15.24\\
     E&\checkmark&\checkmark & &\checkmark & & 16.92 & 13.81 & 25.15 & 17.40 & 20.85 & 18.63\\
     Ours&\checkmark&\checkmark & & & \checkmark & \bf15.82 & \bf11.96 & \bf15.04 & \bf13.43 & \bf13.93 & \bf14.04 \\
  \hline
  \end{tabular}
\end{table*}

\subsubsection{Ablation on Mask Number} 
\label{AblantionK}

We conducted an ablation study concerning the number of masked point clouds ($K$). We set $K$ to 1, 2, 3, and 4 respectively. The results of the per-point $L_2$ Chamfer Distance $\times 10^4$ are presented in Table~\ref{tab:ablationK}. 
The results indicate that more masked input cannot bring any performance improvement. Thus to save on storage and computational expenses, we set $K=1$ for all experiments.

\begin{table}[h!]
 \renewcommand\arraystretch{1.2}
  \caption{Ablation study on the 3DFUTURE dataset regarding the number of masked point clouds $K$, measured in terms of per-point $L_2$ Chamfer Distance$\times10^4$ (lower is better).
  }
  \label{tab:ablationK}
  \centering
  \setlength{\tabcolsep}{5pt}
      \begin{tabular}{c|cccccc}
      
        \hline
         $K$ & Cabinet & Chair & Lamp & Sofa & Table & Avg. \\
        \hline 
        $K = 1$ & 15.82 & 11.98 & 15.04 & 13.43 & 13.93 & 14.04 \\
        $K = 2$ & 15.98 & 12.07 & 15.22 & 13.50 & 14.43 & 14.24 \\
        $K = 3$ & 15.96 & 12.73 & 15.00 & 13.49 & 13.65 & 14.17\\
        $K = 4$ & 15.90 & 12.05 & 15.47 & 13.50 & 14.68 & 14.32 \\
      \hline
    \end{tabular}
    
\end{table}

\subsubsection{Ablation on the Number of Sampled Point Clouds} 
\label{sec:ablationSample}

We conducted an ablation study on the number of sampled point clouds  ${FPS}\left({P}_{fine}^{s}\right)$ in Coarse-to-fine Point Cloud Distillation. The results are presented in Table~\ref{tab:ablationSample}. 
We observe that with the increase in sampled numbers, the performance in most categories decreases to some extent. 
Thus, we sample 256 points to represent the global structure in our method.

\begin{table}[h]
\renewcommand\arraystretch{1.2}
  \caption{Ablation study on the 3DFUTURE dataset regarding the number of sampled point clouds, measured in terms of per-point $L_2$ Chamfer Distance$\times10^4$ (lower is better).
  }
  \label{tab:ablationSample}
  \centering
  \setlength{\tabcolsep}{5pt}
      \begin{tabular}{c|cccccc}
      
        \hline
         Sample Number & Cabinet & Chair & Lamp & Sofa & Table & Avg. \\
        \hline 
        128 & 15.86 & 11.92 & 14.92 & 13.68 & 14.31 & 14.14 \\
        256 & \bf15.82 & 11.98 & 15.04 & \bf13.43 & \bf13.93 & \bf14.04 \\
        512 & 15.87 & \bf11.78 & 15.29 & 13.64 & 14.27 & 14.17 \\
        1024 & 16.18 & 12.26 & 15.06 & 13.54 & 13.94 & 14.20 \\
        2048 & 16.35 & 12.67 & \bf14.70 & 13.56 & 14.03 & 14.26 \\
      \hline
    \end{tabular}
    
\end{table}

\begin{figure*}[htbp]
    \centering
    \includegraphics[width=\textwidth]{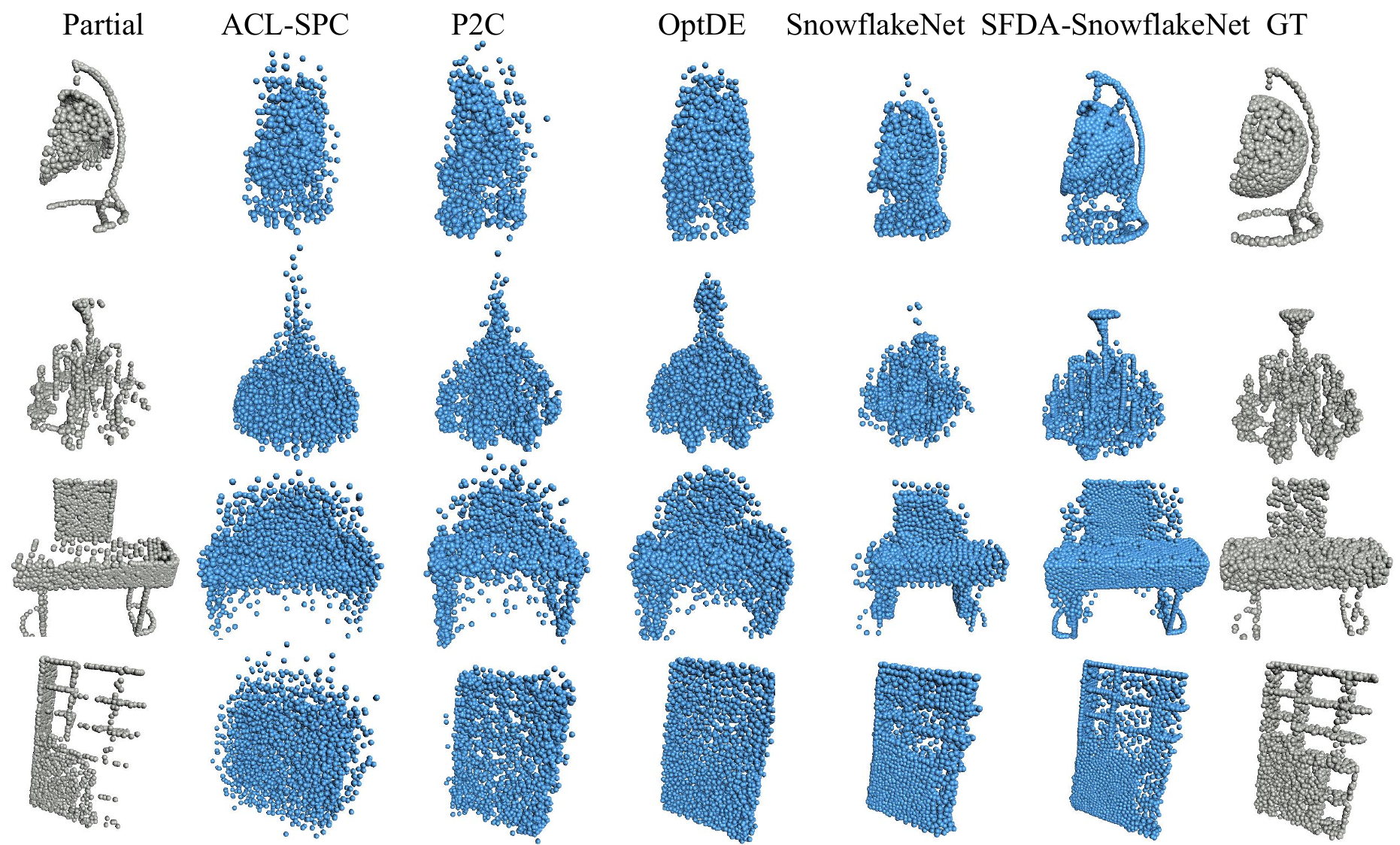}
    \caption{Visualization results of failure cases. PointSFDA may generate redundant
points when the sample contains highly complicated geometric
details. Nevertheless, even in these challenging scenarios, PointSFDA outperforms competitors (ACL-SPC~\cite{ACLSPC}, P2C~\cite{P2C}, OptDE~\cite{OptDE}) by reconstructing more plausible shapes.}
    \label{fig:failure}
\end{figure*}

\subsubsection{Ablation on Mask Strategy} 
\label{AblantionMask}
We also conducted an ablation study regarding mask strategy. The results are presented in Table~\ref{tab:ablationMask}. In the ``Partition'' strategy, we partition the input point cloud $P_{in}$ into 8 equal parts in 3D space and randomly mask one of them each time to obtain masked point clouds. 
In the ``View'' strategy, we select one random point and then mask out a fixed number of points nearest to it. To ensure a fair comparison, the number of masked points is set to $1/8$ of the total number of points. As shown in Table~\ref{tab:ablationMask}, the ``Partition'' strategy exhibits a slight advantage over the ``View'' strategy. Consequently, we chose to employ the "Partition" strategy in our method.

\begin{table}[htbp]
\renewcommand\arraystretch{1.2}
  \caption{Ablation study on the 3DFUTURE dataset regarding the method of 
 masking, measured in terms of per-point $L_2$ Chamfer Distance$\times10^4$ (lower is better).
  }
  \label{tab:ablationMask}
  \centering
  \setlength{\tabcolsep}{5pt}
      \begin{tabular}{c|cccccc}
      
        \hline
         Masking Method & Cabinet & Chair & Lamp & Sofa & Table & Avg. \\
        \hline 
        Partition & \bf15.82 & \bf11.98 & \bf15.04 & 13.43 & 13.93 & \bf14.04 \\
        View & 16.96 & 12.07 & 16.62 & \bf12.97 & \bf13.25 & 14.37 \\
      \hline
    \end{tabular}
    
\end{table}

\label{sec:experiments}

\section{Limitation}
Figure~\ref{fig:failure} contains the failure cases we observed.
We find that a limitation of our method is its tendency to generate redundant points when the sample contains highly complicated geometric details. Despite this limitation, when compared to alternative methods, PointSFDA outperforms in both global structure and preservation of geometric details. 
A potential solution to this problem is to utilize point cloud denoising techniques in our SFDA framework.

\section{Conclusion}
We propose PointSFDA for source-free domain adaptation of point cloud completion. PointSFDA is a general architecture applicable to all point cloud completion networks that generate point clouds in a coarse-to-fine manner. PointSFDA can effectively transfer geometric knowledge from the pre-trained source model to the target model by Coarse-to-fine Point Cloud Distillation. Meanwhile, we introduce partial-mask consistency training to learn the local structure information of the target domain. Extensive experiments demonstrate that PointSFDA can significantly improve the performance of cross-domain point cloud completion.

\bibliographystyle{IEEEtran} 
\bibliography{references} 

\end{document}